\documentclass[twocolumn,superscriptaddress,preprintnumbers,amsmath,10pt,aps,prl,nobibnotes]{revtex4}
\usepackage{graphicx}
\usepackage{amsmath}
\usepackage{amsfonts}
\usepackage{bm}
\usepackage{color}
\usepackage{verbatim}

\newcommand{\Tnm}{\bm{T}}

\begin{document}

\title{Adaptive foveated single-pixel imaging with dynamic super-sampling}

\author{David~B.~Phillips}
\email{david.phillips@glasgow.ac.uk}
\affiliation{SUPA, School of Physics and Astronomy, University of Glasgow, Glasgow, G12 8QQ, UK.}
\author{Ming-Jie~Sun}
\email{mingjie.sun@buaa.edu.cn}
\affiliation{SUPA, School of Physics and Astronomy, University of Glasgow, Glasgow, G12 8QQ, UK.}
\affiliation{Department of Opto-Electronic Engineering, Beihang University, Beijing, 100191, China}
\author{Jonathan~M.~Taylor}
\affiliation{SUPA, School of Physics and Astronomy, University of Glasgow, Glasgow, G12 8QQ, UK.}
\author{Matthew~P.~Edgar}
\affiliation{SUPA, School of Physics and Astronomy, University of Glasgow, Glasgow, G12 8QQ, UK.}
\author{Stephen~M.~Barnett}
\affiliation{SUPA, School of Physics and Astronomy, University of Glasgow, Glasgow, G12 8QQ, UK.}
\author{Graham~G.~Gibson}
\affiliation{SUPA, School of Physics and Astronomy, University of Glasgow, Glasgow, G12 8QQ, UK.}
\author{Miles~J.~Padgett.}
\affiliation{SUPA, School of Physics and Astronomy, University of Glasgow, Glasgow, G12 8QQ, UK.}


\begin{abstract}

As an alternative to conventional multi-pixel cameras, single-pixel cameras enable images to be recorded using a single detector that measures the correlations between the scene and a set of patterns. However, to fully sample a scene in this way requires at least the same number of correlation measurements as there are pixels in the reconstructed image. Therefore single-pixel imaging systems typically exhibit low frame-rates. To mitigate this, a range of compressive sensing techniques have been developed which rely on a priori knowledge of the scene to reconstruct images from an under-sampled set of measurements. In this work we take a different approach and adopt a strategy inspired by the foveated vision systems found in the animal kingdom - a framework that exploits the spatio-temporal redundancy present in many dynamic scenes. In our single-pixel imaging system a high-resolution foveal region follows motion within the scene, but unlike a simple zoom, every frame delivers new spatial information from across the entire field-of-view. Using this approach we demonstrate a four-fold reduction in the time taken to record the detail of rapidly evolving features, whilst simultaneously accumulating detail of more slowly evolving regions over several consecutive frames. This tiered super-sampling technique enables the reconstruction of video streams in which both the resolution and the effective exposure-time spatially vary and adapt dynamically in response to the evolution of the scene. The methods described here can complement existing compressive sensing approaches and may be applied to enhance a variety of computational imagers that rely on sequential correlation measurements.

\end{abstract}
\maketitle

Computational imaging encompasses techniques that image using single-pixel detectors in place of conventional multi-pixel image sensors~\cite{Sen2005,Takhar2006}. This is achieved by encoding spatial information in the temporal dimension~\cite{Goda2009}. Using this strategy, images are reconstructed from a set of sequential measurements, each of which probes a different subset of the spatial information in the scene. This enables imaging in a variety of situations that are challenging or impossible with multi-pixel image sensors~\cite{Hunt2013}. Examples include imaging at wavelengths where multi-pixel image sensors are unavailable, such as in the terahertz band~\cite{Chan2008,Watts2014,Stantchev2016}, 3D ranging~\cite{Kane1987,Howland2011,Howland2013,Sun2016a}, and fluorescence imaging through pre-characterised multimode fibres and scattering media~\cite{Popoff2014,vCivzmar2012,Mahalati2013,Ploschner2015}.

In order to fully sample an unknown scene to a particular resolution, the minimum number of measurements required is equal to the total number of pixels in the reconstructed image. Therefore, doubling the linear resolution increases the required number of measurements by a factor of 4, leading to a corresponding reduction in frame-rate. This trade-off between resolution and frame-rate has led to the development of a range of compressive techniques that aim to use additional prior knowledge or assumptions about a scene to reconstruct images from an under-sampled set of measurements~\cite{Wakin2006,Candes2007,Baraniuk2007,Duarte2008,Katz2009}. 

Despite these challenges, computational imaging approaches also potentially offer new and more flexible imaging modalities. For example, the lack of a fixed Cartesian pixel geometry means it is no longer necessary for the resolution or exposure-time (i.e.\ the time taken to record all the measurements used in the reconstruction of an image) to remain uniform across the field-of-view, or constant from frame to frame~\cite{Abetamann2013,Yu2014,Soldevila2015,Chen2015}.

A variety of animal vision systems successfully employ spatially-variant resolution imaging~\cite{Crescitelli1972,Banks2015}. For example the retina in the vertebrate eye possesses a region of high visual acuity (the {\it fovea centralis}) surrounded by an area of lower resolution (peripheral vision)~\cite{OBRIEN1951}. The key to the widespread success of this form of {\it foveated} vision is in its adaptive nature. Our gaze, which defines the part of the scene that is viewed in high resolution during a period of {\it fixation}, is quickly redirected (in a movement known as a {\it saccade}) towards objects of interest~\cite{Becker1969,Fischer1984}. Unlike a simple zoom, the entire field-of-view is continuously monitored, enabling saccadic movement to be triggered by peripheral stimuli such as motion or pattern recognition~\cite{Morrone1988,Duncan1994,Strasburger2011}. Space-variant vision exploits the temporal redundancy present in many dynamic scenes to reduce the amount of information that must be recorded and processed per frame:
essentially performing lossy compression at the point of data acquisition. This in turn speeds up the frame-rate of such a vision system, and enables us to react to our surroundings more fluidly.

In this work we demonstrate how an adaptive foveated imaging approach can enhance the useful data gathering capacity of a single-pixel computational imaging system. We note that there has already been much interest in mimicking animal imaging systems for image compression and robotic vision~\cite{Bolduc1998,Geisler1998,Stasse2000}, and our work extends this to the constricted bandwidth regimes of single-pixel computational imagers. Here we reduce the number of pixels in each raw frame (thereby increasing the frame-rate) by radially increasing the size of pixels away from a high-resolution foveal region~\cite{Kortum1996,Carles2016}. The position of the fovea within the field-of-view can then be guided by a variety of different visual stimuli detected in previous images~\cite{Yamamoto1996}. 

Furthermore, we also borrow a concept from the {\it compound} eye architecture to increase the resolution of our images in the periphery: the fusion of multiple low resolution frames to synthesise a higher resolution image of the scene (a technique also known as super-sampling or digital super-resolution~\cite{Tanida2001,Park2003,Carles2014}). In this way we rapidly record the details of fast changing or important features in a single frame, whilst simultaneously building up detail of more slowly changing regions over several consecutive frames. We show how this tiered form of digital super-resolution enables the reconstruction of composite images which possess both a spatially-varying resolution and a spatially-varying effective exposure-time, which can be optimised to suit the spatio-temporal properties of the scene. We demonstrate an implementation of our technique with a single-pixel camera, however the method can be applied to enhance the performance of a growing range of computational imaging systems that reconstruct images from a set of sequential measurements.

\section{Foveated single-pixel imaging}

Single-pixel imaging is based on the measurement of the level of correlation between the scene and a series of patterns. The patterns can either be projected onto the scene (known as structured illumination~\cite{Sen2005}, and closely related to the field of computational ghost imaging~\cite{Shapiro2008,Bromberg2009,Phillips2016}), or used to passively mask an image of the scene (structured detection~\cite{Takhar2006}), which is the method we use here.

\begin{figure}[t]
\includegraphics[width=0.85\linewidth]{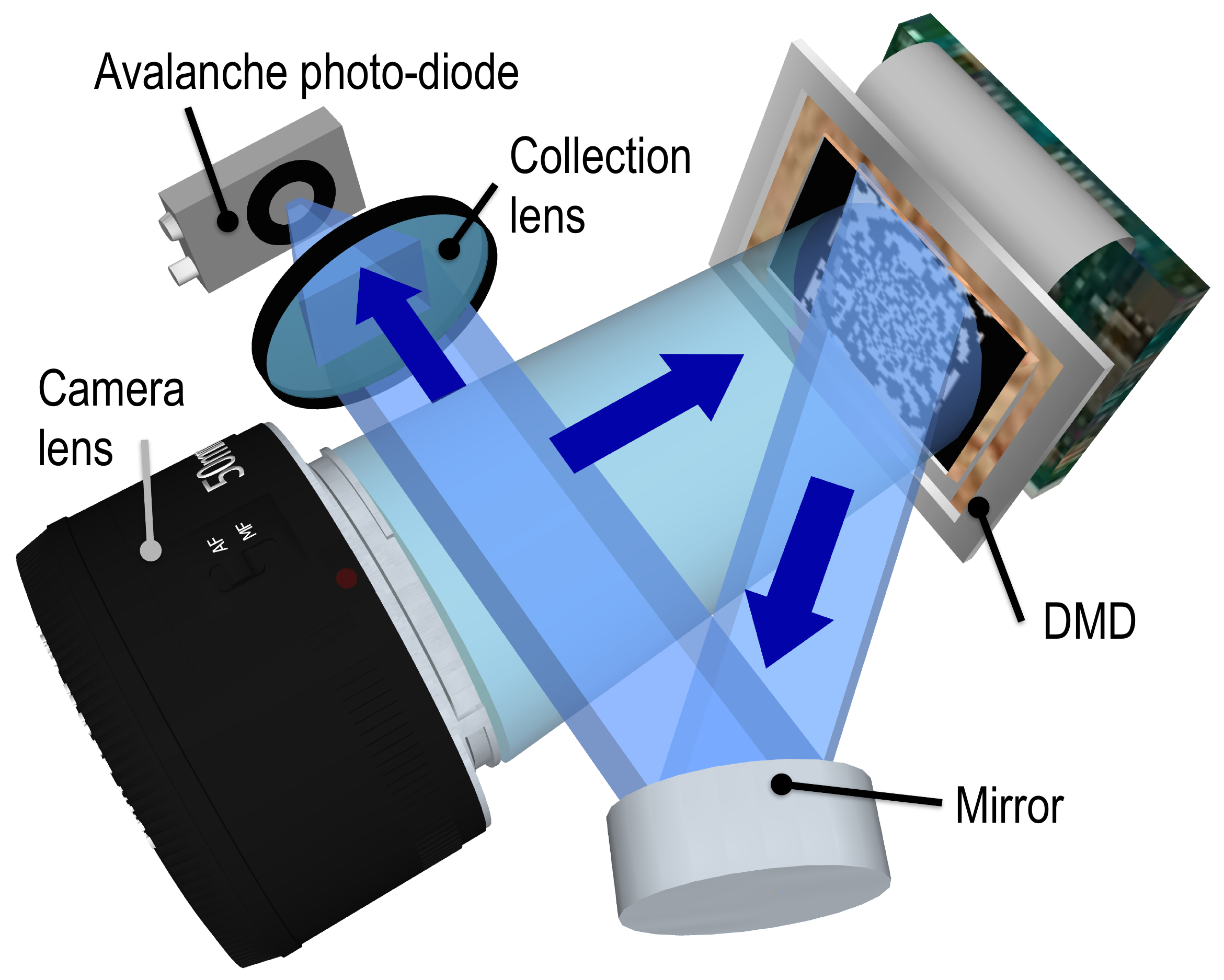}
\caption{{\bf Experimental set-up}. The scene is imaged onto a DMD (Texas Instruments Discovery 7001 with Vialux software) using a Nikon F-mount camera lens (AF NIKKOR 85mm 1:1.8D). The DMD operates as a dynamic mask, reflecting only light from a subset of the pixels to an avalanche photo-diode (APD) (Thorlabs PMM02), which records the total intensity transmitted by each binary masking pattern. The scene is illuminated with a LED torch (Maglite).}
\label{fig0}
\end{figure}

A schematic of the single-pixel camera used in this work is shown in Fig.~\ref{fig0}, based on the design previously demonstrated in refs.~\cite{Edgar2015} and~\cite{Sun2016}. A digital micro-mirror device (DMD) is placed at the image plane of a camera lens, i.e.\ at the same plane where a multi-pixel camera sensor would be placed in a conventional camera. The DMD is used to rapidly mask the image of the scene with a set of binary patterns, and the total amount of light transmitted by each mask is recorded by a photodiode, representing a measurement of the level of correlation of each mask with the scene. Knowledge of the transmitted intensities and the corresponding masks enables reconstruction of the image.

By choosing a set of linearly independent masks, the scene can be critically sampled in an efficient manner using the same number of masks as there are pixels in the reconstructed image. A mask set that is widely-used for single-pixel imaging is formed from the Hadamard basis, which is a set of orthonormal binary functions with elements that take the value of +1 or -1~\cite{Sloane1976,Davis1995,Harwit2012}.
This represents a convenient choice of expansion as when represented on a DMD, each mask transmits light from approximately half of the image pixels, thus maximising the signal at the photodiode.

\begin{figure*}[t]
\includegraphics[width=0.8\linewidth]{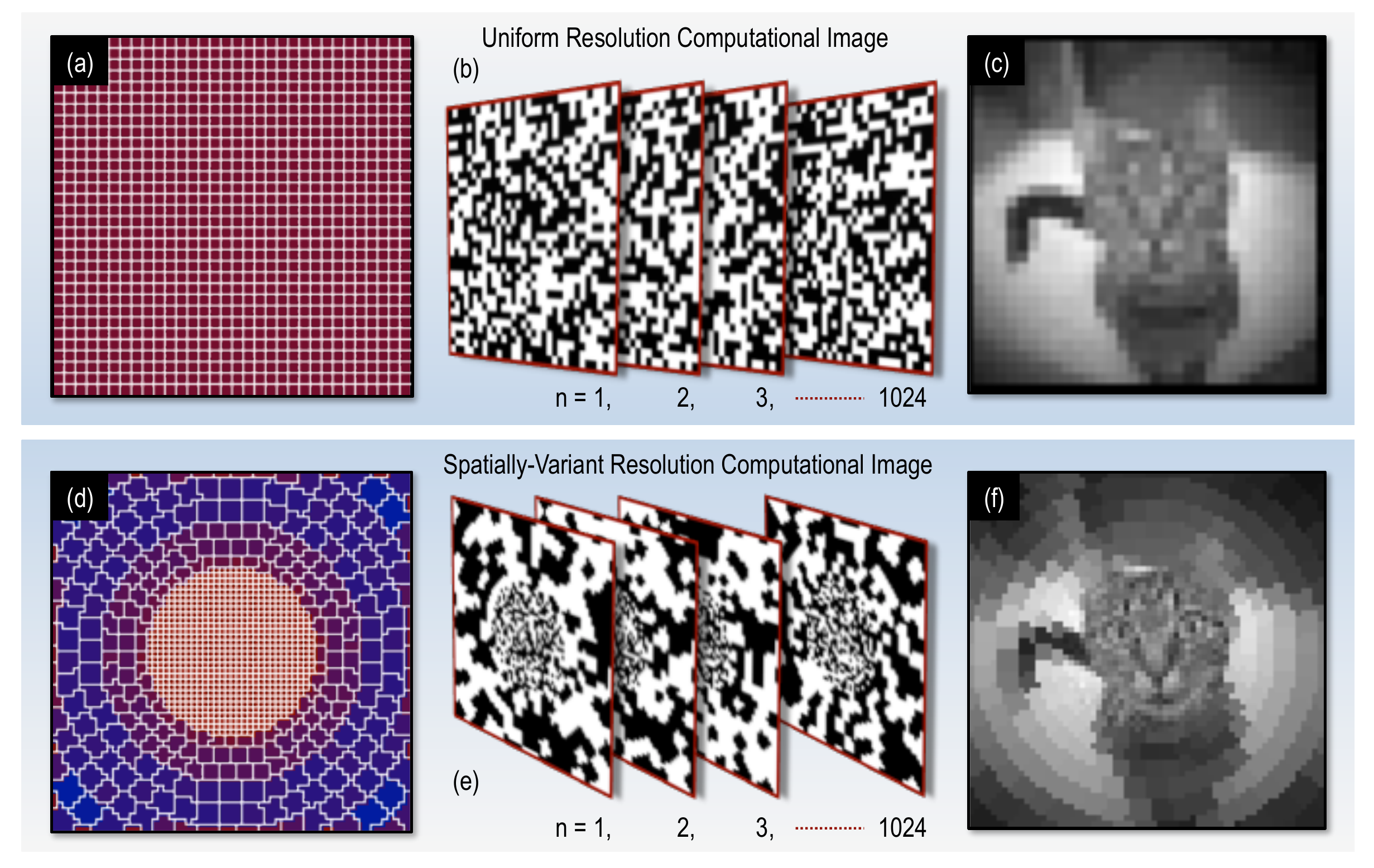}
\caption{{\bf Computational imaging with a spatially-variant resolution}. {\it Top row: uniform resolution}. (a) A uniform $32\! \times \! 32$\,pixel grid with $N$=1024\,pixels. (b) Examples of a complete 1024 Hadamard pattern set (negatives not shown) reformatted onto the 2D uniform grid shown in (a). (c) An image of a picture of a cat recorded experimentally in $\sim$\,0.125\,s, reconstructed from the level of correlation with each of the 1024 masks shown in (b). {\it Bottom row: spatially-variant resolution}. (d) A spatially-variant pixel grid, also containing $N$=1024\,pixels. Within the fovea the pixels follow a Cartesian grid, chosen to avoid aliasing with the underlying Cartesian grid of the DMD at high resolutions. Surrounding the fovea is a peripheral region possessing a cylindrical-polar system of pixels. (e) Examples of the 1024 Hadamard patterns reformatted onto the spatially-variant grid shown in (a). (f) An image of the identical scene to that shown in (c), here reconstructed from correlations with the 1024 spatially-variant masks shown in (e). In the central region of (f) the linear resolution is twice that of the uniform image (c). }
\label{fig1}
\end{figure*}

A uniform resolution $N$ pixel image of the scene (represented here by an $N$ element column vector $\bm{o}_{un}$), can be expressed as a linear sum in our basis of $N$ Hadamard vectors, index $n$ of which is denoted by $\bm{h}_n$:
\begin{equation}
\label{eq:regHad}
\bm{o}_{un} = \frac{1}{N}\sum_{n=1}^N a_n \bm{h}_n,
\end{equation}
where $a_n$ is the level of correlation between the scene $\bm{o}_{un}$ (sampled to the same resolution as the Hadamard patterns) and mask $n$ recorded by the photodiode, i.e. $a_n$ is measured by projecting the scene onto the $n^{\textrm{th}}$ Hadamard mask: $a_n = \bm{h}_n^T\bm{o}_{un}$, which follows from the orthogonality of the Hadamard vectors (i.e.\ $\bm{h}_n^T\bm{h}_m=N\delta_{nm}$). We emphasize that any spatial frequency components in the scene that are above the spatial frequency limit of the uniform pixel grid are lost in this process. Although presented here in 1D vector notation, experimentally a 2D image is recorded, with each 1D vector $\bm{h}_n$ being reshaped onto a uniform 2D grid which is displayed on the DMD, as shown in Figs.~\ref{fig1}(a-b). See Section S1 for more detail.

Figure~\ref{fig1}(c) shows an example of an experimentally reconstructed image of uniform resolution containing $32\! \times \! 32$ pixels. Exploiting a fast DMD (see caption of Fig.~\ref{fig0}) enables the display of $\sim$\,2$\times$$10^4$\,masks/s, resulting in a reconstructed frame-rate of $\sim$\,10\,Hz at this $32\! \times \! 32$ pixel resolution (incorporating two patterns per pixel for differential measurement to improve SNR, see Section S1).
Evidently it is highly desirable to try increase the useful resolution-frame-rate product of such a single-pixel computational imaging system.\\

We use a modification of the technique described above to measure and reconstruct images of non-uniform resolution. In this case the masking patterns displayed on the DMD are created by reformatting each row of the Hadamard matrix into a 2D grid of spatially-variant pixel size, as shown in Figs.~\ref{fig1}(d) and~\ref{fig1}(e). For clarity, we henceforth refer to these non-uniformly sized pixels as {\it cells}. Here the 2D grid has an underlying Cartesian resolution of $M = 64\! \times \! 64 = 4096$ pixels, but contains $N = 1024$ independent cells. Mathematically, this reformatting operation may be expressed as a transformation of the Hadamard vectors to a new set of vectors~$\bm{s}$ using the matrix $\Tnm$ which is a $M\! \times \!  N$ (rows\,$\times$\,columns) binary matrix that stretches a vector of $N$ elements (representing the number of cells) to populate a (larger) vector of $M$ high-resolution pixels: $\bm{s}_n = \Tnm{}\bm{h}_n$.
Similarly to above, we measure the correlation $b_n$ between each pattern $\bm{s}_n$ and the scene $\bm{o}$ (where here $\bm{o}$ is an $M$ element vector representing the scene at uniform resolution equivalent to the highest resolution of patterns $\bm{s}$). Therefore $b_n = \bm{s}_n^T\bm{o}$.

Due to the stretch transformation, the masks $\bm{s}$ are no longer orthogonal. However, the spatially-variant image of the scene, $\bm{o}_{sv}$, can still be efficiently reconstructed using:
\begin{equation}
\label{eq:o-sv-recovery}
\bm{o}_{sv} = \bm{A}^{-1} \frac{1}{N}\sum_{n=1}^N b_n \bm{s}_n.
\end{equation}
Here $\bm{A}$ is an $M\! \times \!  M$ diagonal matrix encoding the area of each pixel in the stretched basis: element $A_{mm}$ is equal to the area of the cell to which high-resolution pixel $m$ belongs. Section S2 gives a detailed derivation of this result. Unlike before, in this case the high spatial frequency cut-off is now spatially-variant across the field-of-view. We also note that the signal-to-noise ratio (SNR) is now also spatially-variant, see Section S3 for more detail. 

Figure~\ref{fig1}(f) shows an experimentally reconstructed spatially-variant resolution image of the same scene as shown in Fig.~\ref{fig1}(c). Although both of the images use the same measurement resource (i.e. each has the same total number of independent pixels, and therefore each requires the same number of mask patterns and effective exposure-time to measure), the linear resolution in the central region of Fig.~\ref{fig1}(f) is twice that of Fig.~\ref{fig1}(c). The detail in the foveal region (the cat's face) is therefore enhanced in Fig.~\ref{fig1}(f) at the expense of lower resolution in the surroundings. 

\begin{figure*}[tt]
\includegraphics[width=\linewidth]{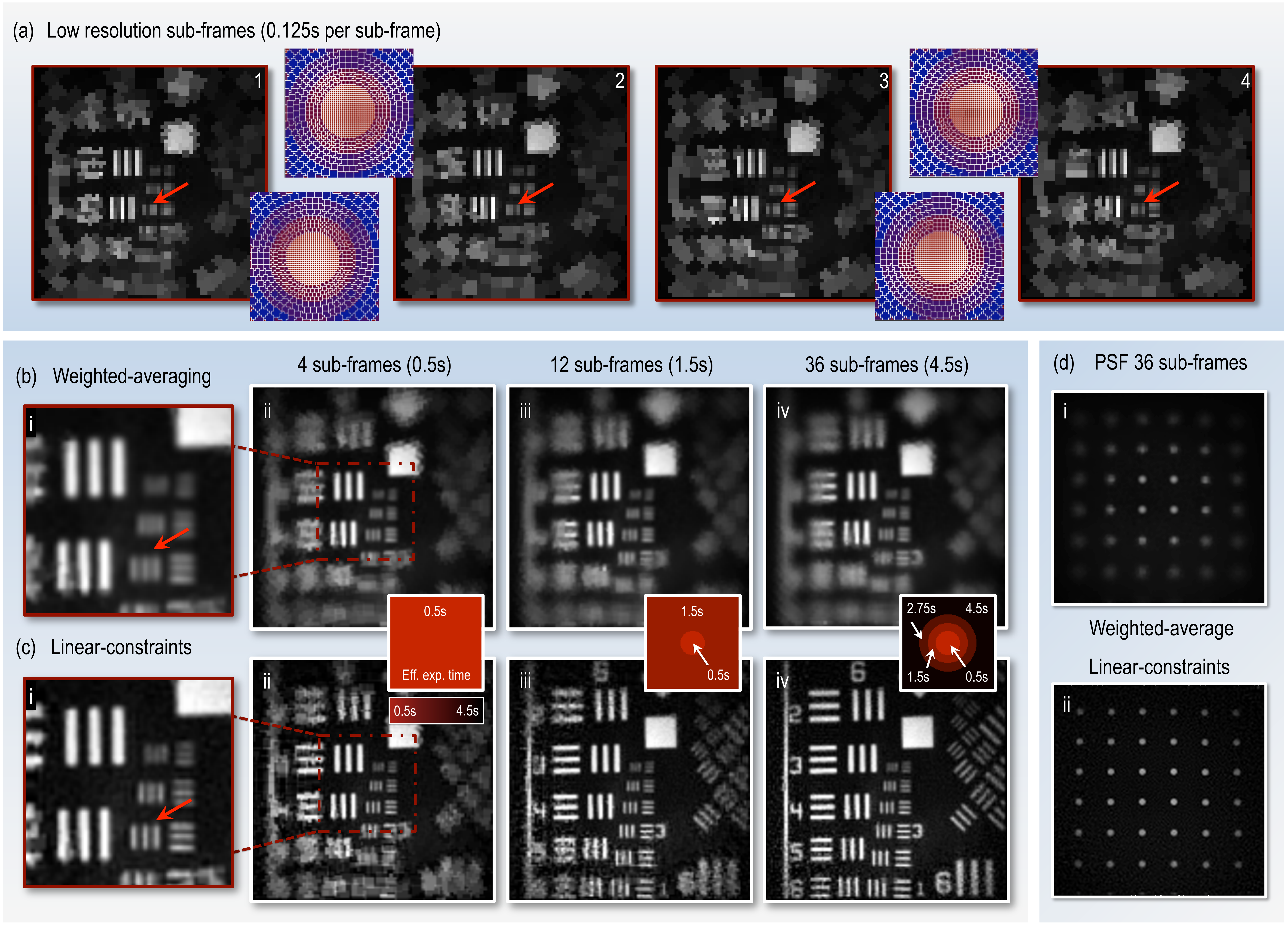}
\caption{{\bf Reconstructing images with a spatially-variant effective exposure-time using digital super-sampling}. All images are reconstructed from experimental data. (a) 4 sub-frames, each with the foveal cells shifted by half a cell in $x$ and/or $y$ with respect to one another~\cite{Sun2016}. The number of cells in each sub-frame $N=1024$. The purple insets show the underlying cell grid in each case. Media~1 shows the changing footprints of the sub-frames in real-time (see S6 for full description of Media 1). (b) Composite images reconstructed from increasing numbers of sub-frames using the weighted-averaging method. (c) Composite images reconstructed from increasing numbers of sub-frames using the linear-constraint method. The number of hr-pixels in the high-resolution composite images is $M = 128\! \times \! 128 = 16384$, although not all of these may be independently recovered, depending upon the number of sub-images combined and the configuration of each sub-image's cells. Insets bridging (b) and (c) colour code the local time taken to perform the measurements used to reconstruct each region within the field-of-view, i.e.\ the spatially-variant {\it effective} exposure-time of the images. The central region only ever uses data from the most recent 4 sub-frames (taking 0.5\,s), whilst the reconstruction of the periphery uses data from sub-frames going progressively further back in time. Media~2 shows a movie of the progressive linear constraint reconstruction (see S6 for full description of Media 2). (d) Reconstructions of a uniform grid of points from 36 sub-frames to compare the PSF of the two reconstruction methods.}
\label{fig2}
\end{figure*}

\section{Spatially-variant digital super-sampling}
If the positions of the pixel boundaries are modified from one frame to the next, then each frame samples a different subset of the spatial information in the scene. Consequently, successive frames are not only capturing information about the temporal variation of the scene, they are also capturing additional, complementary information about the spatial structure of the scene. Therefore, if we know that a local region of the scene has been static during the course of the measurements, we can combine these measurements to recover an image of \emph{enhanced resolution} compared to the reconstruction of an individual frame in isolation. This technique is known as digital super-resolution or super-sampling~\cite{Park2003}. As the pixel geometry of each frame in our single-pixel imaging system is defined by the masking patterns applied to the DMD and used to measure the image, it is straightforward to modify the pixel boundaries from frame to frame as required, and the images are inherently co-registered for digital resolution enhancement. We note that the term digital super-resolution refers to increasing the resolution in imaging systems in which the resolution is limited by the pixel-pitch and not the diffraction limit.

Figure~\ref{fig2} demonstrates how digital super-sampling can be combined with spatially-variant resolution, which leads to reconstructions with different effective exposure-times across the field-of-view. For clarity, we henceforth refer to the raw images (shown in Fig.~\ref{fig2}(a)) as {\it sub-frames} (which contain the non-uniformly sized cells), and uniform Cartesian pixels of the high-resolution composite reconstruction as {\it hr-pixels}. Our DMD can be preloaded with a set of masks to measure up to $\sim$\,36 different sub-frames, each containing 1024 cells with different footprints, which once loaded can be played consecutively in an arbitrary and rapidly switchable order.

Within the fovea where the cells occupy a regular square grid the linear resolution can be doubled by combining 4 sub-frames of overlapping fovea positions. To achieve this the cell footprints are translated by half a cell's width in the $x$ and/or $y$-direction with respect to the other sub-frames (more detail of the relative cell positions is given in~\cite{Sun2016}). Media 1 shows the recording of the sub-frames in real-time (see S6 for full description). The variation in detail within the fovea of each of the 4 sub-frames can be seen in Fig.~\ref{fig2}(a) where the red arrows highlight regions for comparison. 
The boundaries of the lower-resolution peripheral cells are also repositioned in different sub-frames in Fig~\ref{fig2}(a), but since the peripheral cells lie on a less-regular grid with variable sizes, they cannot be shifted by a constant amount. Instead, they are randomly repositioned with each new sub-frame, which is realised by randomising their azimuth and displacing the centre of each fovea by a small amount. Therefore the resolution is increased non-uniformly in the periphery.

Having acquired this information, we are free to choose different reconstruction algorithms to fuse the information from multiple sub-frames to recover an improved estimate of the original scene $\bm{o}'_{sv}$. These algorithms can be built upon different assumptions about the level of motion within the scene, and trade off real-time performance against quality of super-resolution reconstruction. Here we describe and compare two such fusion strategies: \emph{weighted-averaging} and \emph{linear-constraints}.\\ 

\noindent {\bf Weighted-averaging:} In this first strategy we perform a weighted average of multiple sub-frames to reconstruct an image with increased resolution~\cite{Sun2016}. The sub-frames are upscaled by a factor of 2 and co-registered. Then, within the fovea, the 4 most recent sub-frames (Fig.~\ref{fig2}(a)) are averaged with equal weightings, yielding a higher resolution composite image as shown in Fig.~\ref{fig2}(b). Outside the foveal region the sizes of the cells vary, and we choose weighting factors for each sub-frame that are inversely proportional to the area of the corresponding sub-frame cell that the data is taken from, promoting data from cells that have a smaller area and thus a higher local resolution. Weighting in this way incorporates local data from all sub-frames in every composite image hr-pixel, which has the benefit of suppressing noise. However alternate weighting factors are possible, see Section S4 for further discussion.\\

\noindent {\bf Linear-constraints:} Our second reconstruction strategy makes use of all available data in the measurements. The reconstructed intensity value of a single sub-frame cell represents an algebraic constraint on the sum of the group of hr-pixels that are members of that cell region. Successive sub-frames are acquired with variations on the cell boundaries, which changes the group of hr-pixels corresponding to each sub-frame cell. As long as a local region of the image has remained static during acquisition of multiple sub-frames, these constraints can then be combined together into one self-consistent system of linear equations, which we solve to recover an improved estimate for the intensity value of each hr-pixel in the composite reconstruction, $\bm{o}'_{sv}$. The linear constraints reconstruction is equivalent to a deconvolution of the weighted-averaging reconstruction with the appropriate spatially varying PSF. See Section S5 for more details.\\

Media~2  and Figures~\ref{fig2}(b) and~\ref{fig2}(c) compare reconstructions using these two alternative methods, as increasing numbers of sub-frames are used (see S6 for full description of Media~2) For weighted-averaging, the foveal region reaches a maximum resolution upon combination of the 4 most recent sub-frames with overlapping fovea, and further increasing the number of sub-frames averaged in the periphery smooths the reconstruction but does little to improve its resolution. With the linear-constraint method, the maximum resolution in the foveal region is also reached after only 4 sub-frames, but the point spread function is sharper, and hence high spatial frequencies are reproduced more faithfully. Furthermore, in the peripheral region as larger numbers of sub-frames are fused into the reconstruction the resolution continues to improve. 

Thus our tiered imaging system captures the detail of the central region of the scene at a frame-rate of 8\,Hz, with resolution-doubled images simultaneously delivered at a frame-rate of 2\,Hz. The weighted-average method offers the same frame-rates in the periphery, but with a space-variant broadening of the point spread function (PSF), and hence reduced resolution (see Figure~\ref{fig2}(d)i). For static regions of the scene, the linear-constraint method offers a means to further trade frame-rate for resolution, enabling us to obtain an almost uniform high resolution across the scene (see Figure~\ref{fig2}(d)ii) after fusing data from 36 sub-frames in the periphery. For comparison, uniformly imaging the entire field-of-view at the higher resolution of the composite reconstruction ($128\! \times \! 128$\,hr-pixels) would lower the global frame-rate to 0.5\,Hz. Therefore, in analogy to the resolution trade-off made in an individual sub-frame, using composite image formation with the linear-constraint method we are able to trade a higher frame-rate in the centre for a lower frame-rate at the periphery.

\begin{figure*}[t]
\includegraphics[width=\linewidth]{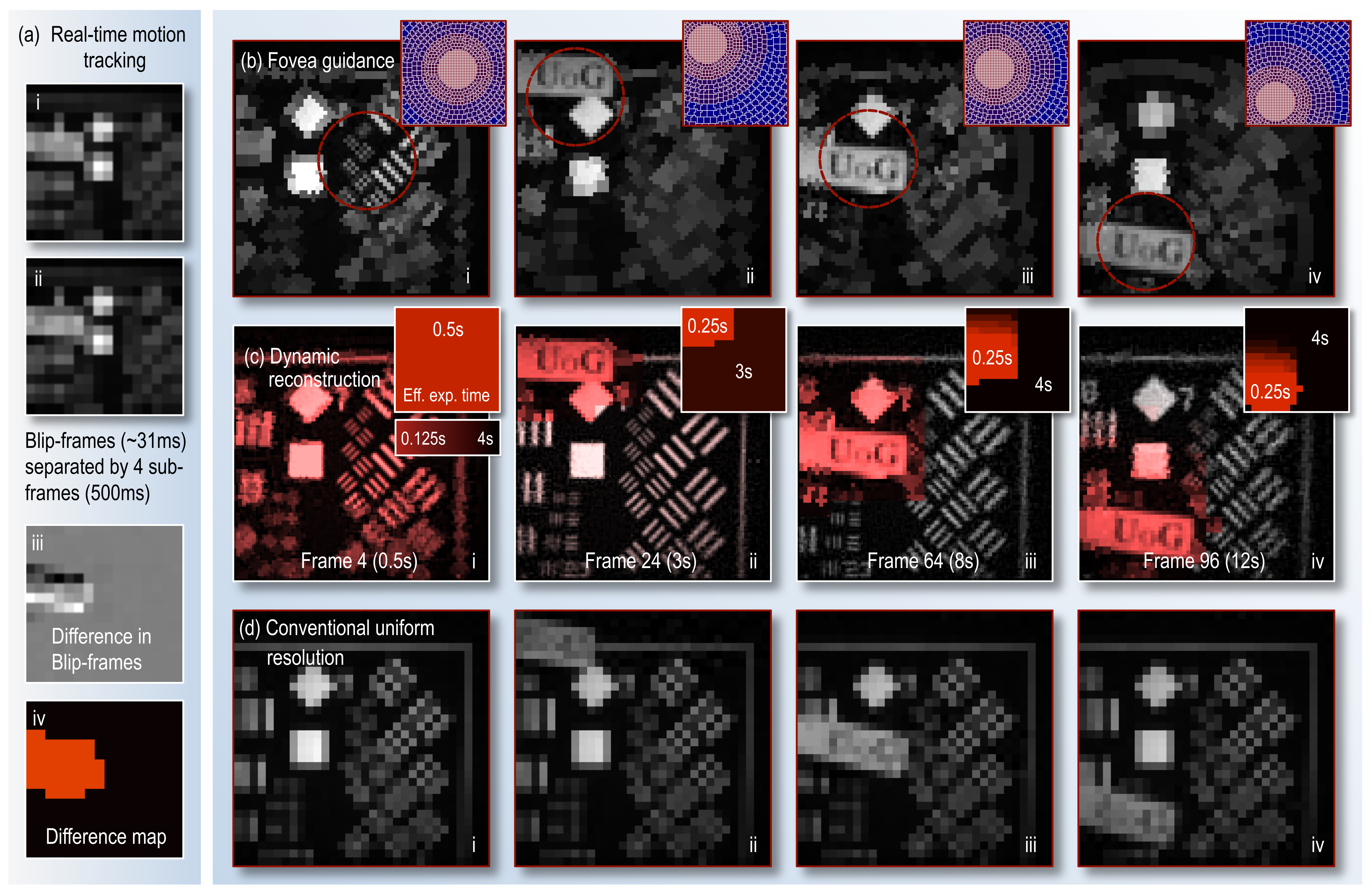}
\caption{{\bf Fovea guidance by motion tracking}. (a, i-ii) Low resolution blip-frames, interlaced between 4 sub-frames. The difference between consecutive blip-frames reveals regions that have changed (iii). A binary difference map (iv) is then constructed by thresholding the modulus of (iii) and then implementing a convex hull operation on the thresholded region to fill any gaps. Finally a dilate operation expands the highlighted area in every direction to ensure it is large enough to accommodate the moving object. This analysis is performed in real-time and so the following sub-frames have a fovea located on the region of the scene that has changed. (b) Frame exerts from Media 3 showing examples of sub-frames (each recorded in 0.125\,s) guided using blip-frame analysis to detect motion (fovea location updated at 2\,Hz). The purple insets show the space-variant cell grid of each sub-frame. (c) Frame exerts from Media 4 showing the reconstructed (using linear constraints) video stream of the scene also capture the static parts of the scene at higher resolution. Here difference map stacks (shown as insets) have been used to estimate how recently different regions of the scene have changed, guiding how many sub-frames can contribute data to different parts of the reconstruction. This represents an effective exposure-time that varies across the field-of-view. Here the maximum exposure-time has been set to 4\,s (i.e.\  all data in the reconstruction is refreshed at most after 4\,s), and the effective exposure-time has also been colour coded into the red plane of the reconstructed images. (d) Conventional uniform resolution computational images of a similar scene for comparison. These use the same measurement resource as (b) and (c).
}
\label{fig3}
\end{figure*}

The improvement in resolution with the linear-constraint method comes at the expense of reconstruction speed. The weighted-averaging technique is fast to compute (scaling as O($N$)), and so can easily be performed in real-time at well above video-rates for the resolutions presented here. In contrast, the linear-constraint method involves finding the least-squares solution to a set simultaneous equations (in our method scaling as O($N^3$)). Here this reconstruction was carried out in post processing (see Section S5 for details), however the use of graphics processors and efficient matrix manipulation could potentially make this problem tractable in real-time for the resolutions demonstrated here~\cite{Murray-Smith2005}.

In the next Section we show how the data gathering capacity of our imaging system can be further improved by dynamically repositioning the fovea within the field-of-view in response to recent measurements, and accounting for parts of the scene which are moving in the reconstruction algorithms.

\section{Fovea gaze control}
As we have described above, the position of the fovea in each sub-frame is determined by displaying a particular sub-set of the patterns that have been preloaded on the DMD. Therefore, mimicking the saccadic movement of animal vision, using real-time feedback control the fovea can be rapidly repositioned in response to cues from previous images, for example to follow motion of objects within the field-of-view or move to areas anticipated to contain high levels of detail.\\

\noindent {\bf Motion tracking:} A range of image analysis techniques exist to estimate motion in dynamic scenes, most involving some form of comparison between two consecutive frames~\cite{Aggarwal1997,Aggarwal1988}. However, image comparison becomes more complicated if the pixel footprints change from one image to the next. Therefore, we have two competing requirements: changing the locations of pixel boundaries is advantageous as it enables digital resolution enhancement of static parts of the scene (as demonstrated in Fig.~\ref{fig2}(c)), yet determining which parts of the scene are in motion is easier if pixel boundaries remain constant between consecutive frames.

In order to balance these requirements, we vary the cell boundaries of consecutive space-variant resolution sub-frames as described above, but also interlace these frames with a short-exposure frame of uniform low resolution (for clarity henceforth referred to as a {\it blip-frame}). The pixel boundaries of the blip-frames never change, and we use comparison of consecutive blip-frames to detect scene motion, as shown in Fig.~\ref{fig3}(a). 
We select relative resolutions for the sub-frames (1024 cells) and blip-frames (16\,$\times$\,16 uniform pixels) and the interlacing frequency (2\,Hz), so as to minimize the impact of the blip-frames on the overall frame-rate.

In the examples here interlacing with a blip-frame reduces the average frame-rate by only $\sim$\,7\,$\%$. Alternatively, to avoid the use of blip-frames we could reconstruct pairs of sub-frames with identical pixel footprints and look for changes between these to track motion. However this strategy would reduce the super-sampling rate by a factor of two.

\begin{figure*}[t]
\includegraphics[width=0.95\linewidth]{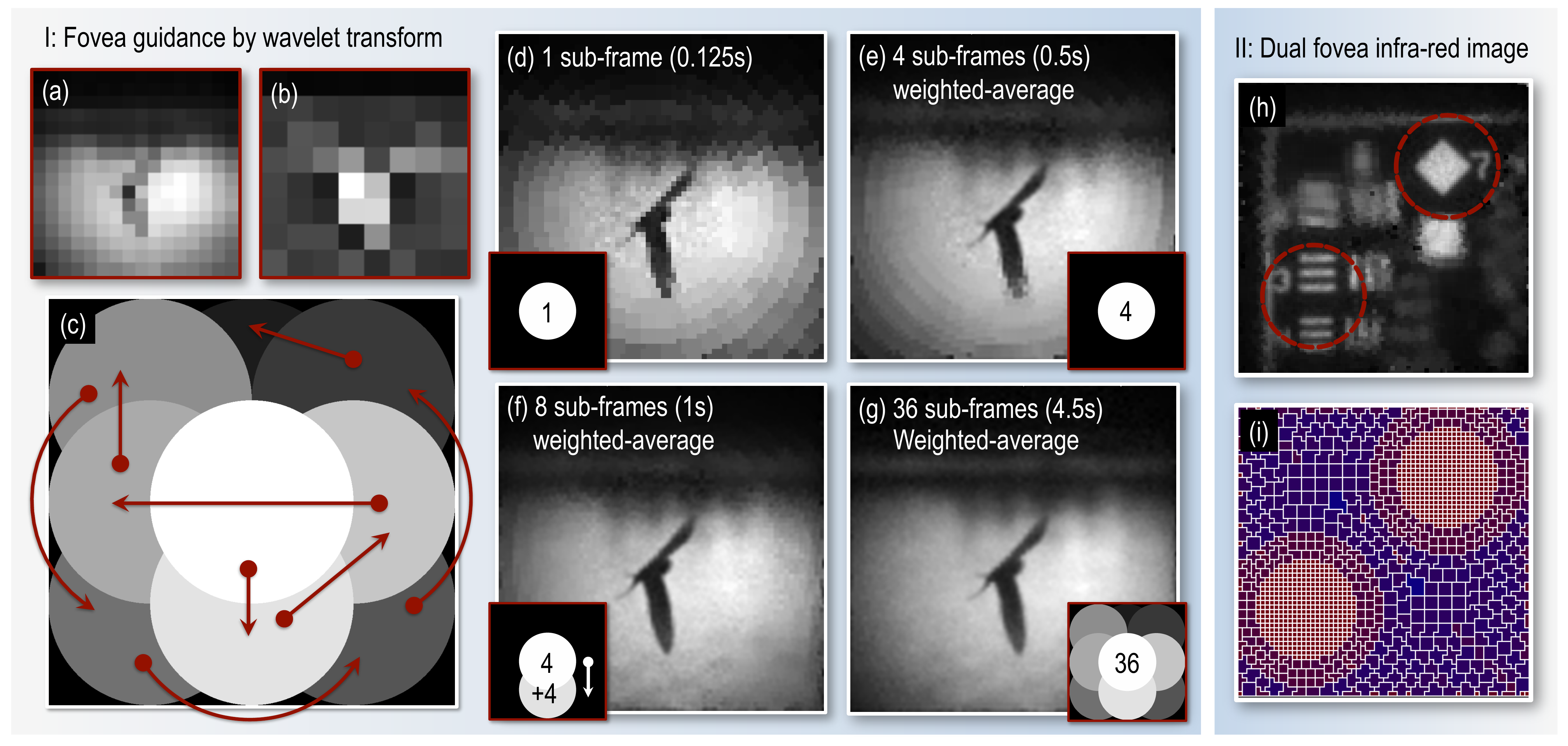}
\caption{{\bf I Fovea guidance by wavelet transform}. The fovea trajectory is determined by first measuring a blip-frame (a). A single tier Haar wavelet transform is then performed on (a). This consists of 4 quadrants: one is a coarse representation of the original image (scaled down by a factor of two). The other three quadrants contain information about the contrast of horizontal, vertical and diagonal edges on the scale of the original images pixels. Regions containing high contrast on the scale of the blip-frame pixels return high values in the wavelet-transform. (b) shows the quadratic sum of the three quadrants containing edge information, which is then used to determine the fovea trajectory. This is achieved by consecutively finding the nearest (thus far unused) fovea locations to the regions of the image containing the highest (thus far un-sampled) contrast. (c) shows a map of the fovea trajectory within the field-of-view. Brighter regions indicate areas that the fovea visits earlier. Arrows show the trajectory of the fovea. (d-g) show image reconstructions after sampling the scene with various numbers of sub-frames and fovea positions. In this example, the fovea trajectory determined by the wavelet transform samples the majority of the detail in the scene after 8 sub-frames. This is 50\% of the time required to sample the entire field-of-view at the same resolution as the centre has been sampled here. {\bf II Dual fovea infra-red image}. Here the APD was replaced with a InGaAs detector (Thorlabs PDA20CS InGaAs, 800-1800\,nm) and the scene illuminated with a heat lamp. (h) shows the weighted-average of 4 sub-frames (1368 cells per sub-frame, frame-rate 6\,Hz), each possessing 2 fovea. (i) the cell grid of one of the sub-frames.}
\label{fig4}
\end{figure*}

Media 3 and Figure~\ref{fig3}(b) shows how motion tracking is used to guide the position of the fovea in real-time (see S6 for full description of Media 3). Here the fovea follows a moving sign containing detail in the form of the letters `UoG', as it is manually swept in front of the camera. The sub-frame frame-rate is 8\,Hz, and in-between every blip-frame we incorporate a `fixation phase', where the fovea stays in the same area but performs 4 sub-frame digital super-sampling as described above. This strategy captures enough information at that location to double the linear resolution within the fovea should the scene remain static. We also inject a stochastic element into the fovea movement: for a randomly chosen fraction $p$ of the sub-frames, the fovea is positioned in a random location not recently accessed, where here p\,$\sim$\,20\%. This ensures that all of the sub-frames are at least intermittently sampled, improving the quality of the longer exposure reconstruction of static parts of the scene.

In addition to guiding the location of the fovea to fast changing parts of the scene, the blip-frames also enable the construction of a dynamic map estimating how recently different regions of the scene last changed.  Composite higher resolution frames can then be reconstructed using stacks of difference maps to determine the local effective exposure-time across the field-of-view (i.e.\ how many sub-frames can contribute data to each region of the reconstruction). Using scene motion estimation to drive both the fovea movement and to build composite images results in a dynamic video reconstruction which can possess significantly enhanced detail in comparison with conventional uniform resolution imaging. This is demonstrated in Media 4 and by comparing Fig.~\ref{fig3}(c) to~\ref{fig3}(d) (see S6 for full description of Media 4). Figure~\ref{fig3}(c) shows examples of composite frame reconstructions using foveated imaging and difference map stacks. The local effective exposure-time has been colour coded into the red channel of the image, highlighting how it changes as the scene evolves. Examples of the stacked difference maps (which also represent the effective exposure-time of frames in the reconstruction) are shown as insets. Figure.~\ref{fig3}(d) shows conventional uniform resolution computational images of a similar scene under the same measurement resource. Here all image data is refreshed at a frame-rate of 8\,Hz, however unlike Fig.~\ref{fig3}(c), the resolution is never high enough to capture detail of the lettering or of the higher resolution parts of the calibration grids. \\

\noindent {\bf Detail estimation:} Depending upon the nature of a dynamic scene, the entire field-of-view may sometimes be temporarily static over the course of several sub-frames. However, beginning to record a uniform high-resolution image at this point is not necessarily the optimum strategy: such a measurement may be quickly interrupted because we have no knowledge of how long the scene will remain static. Therefore it is prudent to attempt to measure the most detail-rich parts of the image first.

In our system we aim to achieve this by performing a single tier Haar wavelet transform on the blip-frame, which yields information about the location of edges in the image, and hence regions of fine detail (refs.~\cite{Burrus1997} and~\cite{Abetamann2013} provide a detailed description of the Haar transform). We use this to calculate a fovea trajectory that samples regions weighing most heavily in the wavelet transformation first, as shown in Fig.~\ref{fig4}.

\noindent {\bf Manual control:} Complementing the automated fovea guidance techniques described above, we have also implemented a manual fovea control system, where a user can click on the part of the scene they wish to view in high-resolution. Other forms of manual control could also be envisaged. For example, control by a single operator could be implemented by measuring eye movements using a gaze tracker, and placing the high-resolution fovea wherever the operator looked. In this case by scaling the resolution profile of the patterns to match the radial visual acuity profile of the eye, the scene could appear to the user to be rendered practically seamlessly in uniformly high-resolution~\cite{Kortum1996}.

\section{Discussion and Conclusions}
In this work we have demonstrated that the data gathering capacity of a single-pixel computational imaging system can be enhanced by mimicking the adaptive foveated vision that is widespread in the animal kingdom. Unlike a simple zoom, in our system every frame delivers new spatial information from across the entire field-of-view, and so this framework rapidly records the detail of fast changing features, while simultaneously accumulating enhanced detail of more slowly changing regions over several consecutive frames. This tiered super-sampling approach enables the reconstruction of video streams where both the resolution and the effective exposure-time vary spatially and adapt dynamically in response to the evolution of the scene.

Unlike many compressive sensing algorithms, our foveated imaging strategy does not require on a priori knowledge of the basis in which the image can be sparsely represented \cite{Wakin2006}. Instead we rely on the assumption that only some regions within the field-of-view will change from frame-to-frame. For many dynamic scenes this is a reasonable assumption, and one that animal vision systems have evolved to incorporate. Our foveated imaging system could potentially be further enhanced if used in conjunction with compressive sensing algorithms, both at the sampling stage (by concentrating measurements in an under-sampled set towards the most important regions of the scene) and reconstruction stage (by incorporating any additional a priori knowledge of the scene to improve accuracy and reduce noise in the composite images). We also note that our composite reconstruction technique is similar in concept to the strategy used in some forms of video compression which also rely on estimation of how recently local regions of the scene have changed~\cite{LeGall1991,Richardson2004}.

We have demonstrated our system at visible wavelengths, however the technique is of course not limited to the visible. For example Fig.~\ref{fig4}(h-i) shows a short wave infra-red (SWIR) image recorded in the wavelength range of 800-1800\,nm, through a piece of perspex opaque to visible light~\cite{Edgar2015}. This is realised by exchanging the avalanche photo-diode with a SWIR sensitive diode, and illuminating with a heat lamp. In addition, Fig.~\ref{fig4}(h-i) also highlights that the number of independently operating fovea can be increased should the scene demand it~\cite{Chen2015}.

In what types of imaging systems might these approaches be most beneficial in the future? The techniques described here may be applied to any form of computational imager performing reconstructions from a set of sequentially made correlation measurements. Despite the challenges of low frame-rates (or low SNR for equivalent frame-rates) exhibited by single-pixel techniques in comparison with conventional multi-pixel image sensors, there are a growing number of situations where cameras cannot easily be used and single-pixel techniques prove highly desirable. For example, recently it has been shown that a form of single-pixel imaging provides a powerful method to transmit image data of a fluorescent scene through pre-calibrated scattering media (such as diffusers or multimode fibres)~\cite{vCivzmar2012,Mahalati2013,Ploschner2015}. Single-pixel techniques also make it possible to image at wavelengths where single-pixel detectors are available, but multi-pixel image sensors are not~\cite{Chan2008,Watts2014,Stantchev2016}. In all such systems there is a trade-off between resolution and frame-rate, and our work demonstrates a flexible means to adaptively optimise this trade-off to suit the nature of the dynamic scene being imaged.

Ultimately, beyond specific technical challenges, the performance of an adaptive foveated computational imaging system will be determined by the sophistication of the algorithms driving the way the scene is sampled. Here we have demonstrated motion tracking using a relatively simplistic algorithm, however the fields of machine and computer vision offer a wealth of more advanced approaches, such as motion-flow algorithms, intelligent pattern recognition and machine learning~\cite{Bishop2006,Rasmussen2006,Sankaranarayanan2015,Li2016}.
The performance of future computational imaging systems can be enhanced by deploying the spatially-variant sampling and reconstruction strategies we have demonstrated here, in partnership with sophisticated image analysis techniques designed to accommodate a variety of real-world situations.

\section{Author contributions}
DBP conceived the concept of the project and developed the idea with support from M-JS, MJP, SB and MPE. M-JS, MPE and GG built the single-pixel camera system. DBP and M-JS developed the imaging software and performed the experiments. DBP developed the reconstruction software and analysed the results with support from JT. SB, JT and DBP derived the theoretical description of the imaging process. DBP and JT wrote the manuscript with support from MJP and input from all other authors.

\section{Acknowledgements}
DBP thanks the Royal Academy of Engineering for support. MJP acknowledges financial support from UK Quantum Technology Hub in Quantum Enhanced Imaging (Grant No. EP/M01326X/1), the Wolfson foundation and the Royal Society. M-JS acknowledges support from the National Natural Foundation of China (Grant No. 61307021) and China Scholarship Council (Grant No. 201306025016). DBP thanks Joy Hollamby and Kevin Mitchell for providing photos to image.\\

Correspondence and requests for materials should be sent to DBP (email: david.phillips@glasgow.ac.uk) or \mbox{M-JS} (email: mingle.sun@buaa.edu.cn).


\section{Supplementary information}

\noindent{\bf 1. Correlation measurements in the Hadamard basis}.
A Hadamard matrix is defined as an $N\! \times \!  N$ matrix with elements that take the values of +1 or -1, and with rows that are orthogonal to one another. The Supplementary Information of reference~\cite{Stantchev2016}, and the references therein give an excellent description of the generation and use of the Hadamard matrices.

In summary a 2D uniform resolution mask of index $n$ is formed by reformatting row $n$ of the Hadamard matrix into a uniform 2D grid, as shown in Figs.~\ref{fig1}(a-b). However, our experimental implementation uses a DMD that can represent masks that transmit (mirrors `on') or block (mirrors `off') intensity regions within the image. This corresponds to masks consisting of +1 (transmitted light) and 0 (blocked light), but not the -1 required by the Hadamard matrix. This problem is circumvented by first displaying a `positive' pattern of +1s and 0s (in place of the -1s) followed by the `negative' of this pattern (i.e.\ where the positions of 1s and 0s have been swapped). The desired Hadamard encoding matrix can then be emulated by subtraction of the intensity transmitted by the negative pattern from the positive pattern. In addition, displaying the negative mask immediately after the positive mask also acts to cancel out some of the noise due to low frequency fluctuations in the ambient illumination, a technique analogous to differential ghost imaging~\cite{Ferri2010}.\\

\noindent{\bf 2. Reconstruction of space-variant resolution sub-frames}.
Here we derive and discuss in further detail Equation~\ref{eq:o-sv-recovery} of the main text. We will make use of the transformation matrix $\Tnm$, which maps from $N$-element cell space to the larger $M$-element hr-pixel space. $\Tnm$ is an $M\! \times \! N$ binary matrix where the locations of the `ones' in column $n$ denote the hr-pixels that belong to cell $n$.

With the help of this matrix, we can define a new basis~$\bm{s}_n$, formed by ``stretching'' the Hadamard vectors $\bm{h}_n$ to conform to our nonuniform pixel grid:
\begin{equation}
\label{eq:sn_definition}
\bm{s}_n =  \Tnm{} \bm{h}_n,
\end{equation}
These are the raw patterns that we will measure using our DMD.
It is important to note that, in contrast to conventional computational imaging using Hadamard matrices on a regular grid, the vectors $\bm{s}_n$ are not orthogonal in hr-pixel space. However, after some matrix algebra it can be shown that there exists a dual basis $\tilde{\bm{s}}_m=\bm{A}^{-1} \bm{s}_m$ forming a biorthogonal set with $\bm{s}_n$, i.e. the following relation holds: 
$$
\bm{s}_n^T \tilde{\bm{s}}_m = \bm{s}_n^T \bm{A}^{-1} \bm{s}_m = N\delta_{mn}.
$$
Here $\bm{A}$ is an $M\! \times \!  M$ diagonal matrix such that $A_{mm}$ is equal to the area of the cell to which hr-pixel $m$ belongs.

The existence of this biorthogonality relationship makes it helpful to represent our high-resolution object $\bm{o}$ in the dual basis $\tilde{\bm{s}}_m$ as follows:
\begin{equation}
\label{eq:o-expansion}
\bm{o} = \frac{1}{N} \sum_{m=1}^N b_m \, \bm{A}^{-1} \, \bm{s}_m + \bm{\epsilon},
\end{equation}
where $\bm{\epsilon}$ represents those high-spatial-frequency components that are orthogonal to all $\bm{s}_n$ (i.e. $\bm{s}_n^T \bm{\epsilon}=0$ for all $n$), and hence that the imaging system described here is not sensitive to.

Then, by projecting $\bm{o}$ onto our basis set $\bm{s}_n$ and expanding, we can show that $\bm{s}_n^T \, \bm{o}=b_n$ as follows: 
\begin{eqnarray}
\bm{s}_n^T \, \bm{o} &=&  \bm{s}_n^T\frac{1}{N} \sum_{m=1}^N b_m \, \bm{A}^{-1} \, \bm{s}_m + \bm{s}_n^T\bm{\epsilon} \nonumber \\
&=& \frac{1}{N}\sum_{m=1}^N b_m \, \bm{s}_n^T \, \bm{A}^{-1} \, \bm{s}_m = b_n. \nonumber
\end{eqnarray}
These measurements can then be used to derive our estimate $\bm{o}_{sv}$ of the object (Equation~\ref{eq:o-sv-recovery} in the main text):
\begin{equation}
\bm{o}_{sv} = \bm{A}^{-1} \frac{1}{N}\sum_{n=1}^N b_n \bm{s}_n.\nonumber
\end{equation}
However, if we substitute (\ref{eq:sn_definition}) into this equation then we see that the same reconstruction in fact be computed more efficiently in cell space:
\begin{equation}
\label{eq:o-sv-recovery-hadamard}
\bm{o}_{sv} = \frac{1}{N}\sum_n b_n \bm{A}^{-1}\Tnm{}  \bm{h}_n = \bm{A}^{-1} \Tnm{} \frac{1}{N}\sum_n b_n \bm{h}_n.
\end{equation}
Although this matrix equation is slightly more long-winded than (\ref{eq:o-sv-recovery}), it shows that the reconstruction can be performed in the (lower-dimensional) cell space, and then the final result remapped just once onto the uniform hr-pixel grid.\\

\noindent {\bf 3. Signal-to-noise ratio in space-variant resolution sub-frames.}
In passive single-pixel imaging techniques, the signal-to-noise ratio (SNR) scales approximately in inverse proportion to the square of the linear resolution, for a given constant exposure-time. Following from this observation, in our spatially-variant resolution imaging system the SNR across each individual sub-frame is also spatially-variant, with the higher resolution regions being most sensitive to noise. Therefore the local SNR scales in inverse proportion to the square of the local linear resolution. The weighted-averaging method does go some way towards improving the SNR, as is discussed in more detail in~\cite{Sun2016}.

However, our space-variant imaging system does offer a reduction in the noise caused by motion blur. When a scene changes {\it during} the measurement of a computational image using Hadamard patterns (uniform or spatially-variant), the reconstruction not only exhibits conventional motion blur, but also a splash of noisy pixels across the field-of-view. This {\it pattern multiplexing} noise is due to scene movement causing inconsistencies in the measured weights of each pattern. By lowering the resolution in regions of the scene deemed static, our foveated imaging system reduces the amount of time required to image a moving part of the scene to a given resolution (in the examples here, by a factor of 4), therefore reducing both conventional motion blur and pattern multiplexing noise.\\

\noindent {\bf 4. Weighted-averaging image fusion}. Each hr-pixel is formed by weighting the contribution of data from each sub-frame in inverse proportion to the area of the corresponding sub-frame cell that the data is taken from. 
Therefore the intensity of pixel $i,j$ in the weighted mean composite image, $o_{wm}(i,j)$, is given by:
\begin{equation}\label{eq:weightedMean}
o_{wm}(i,j) =  \frac{1}{B(i,j)}\sum_{k}\frac{o^{(k)}(i,j)}{A^{(k)}(i,j)},
\end{equation}
where $k$ indexes the sub-frames used to calculate the composite image, $o^{(k)}(i,j)$ is the value of pixel $i,j$ in space-variant sub-frame $k$, and  $A^{(k)}(i,j)$ is the area of the cell that pixel $i,j$ belongs to.  $B(i,j)= \sum_{k}(A^{(k)})^{-1}(i,j)$ serves to normalise the sum. Consistent with our earlier vector notation, we can equivalently write:
\begin{equation}
\label{eq:weightedMeanVectorNotation}
\bm{o}_{wm} =  \bm{B}^{-1}\sum_k (\bm{A}^{(k)})^{-1} \, \bm{o}^{(k)}, 
\end{equation}
where  $\bm{B} = \sum_k (\bm{A}^{(k)})^{-1}$.
Equations~\ref{eq:weightedMean} and~\ref{eq:weightedMeanVectorNotation} therefore specify an equal weighting of sub-frames within the fovea (where the pixels are all of the same size), and in the peripheral region promotes data from pixels that have a smaller area and thus a higher local resolution. This strategy incorporates local data from all sub-frames in every composite image pixel, which has the benefit of suppressing noise.

We note that a variety of other weightings may also be applied. Other examples that we investigated include using only data from the sub-frame with the highest resolution pixel (with equal weighting given in regions where sub-frames have the same sized pixels), and the weighting of more recent measurements more prominently. This second weighting strategy can be applied if some parts of the scene are expected to change throughout the measurement. The weighting choice depends upon the distribution of pixel areas in the sub-frames, the noise levels in the measurement, and the expected level of scene motion.\\

\noindent {\bf 5. Linear-constraints image fusion}. For our linear-constraint algorithm, we fuse information from multiple sub-frames by forming a system of linear equations representing constraints on the high-resolution reconstructed image. The problem can be expressed as:
\begin{equation}
\label{eq:linear-system}
\left(       \begin{array}{c} (\Tnm{}^{(1)})^T \\ (\Tnm{}^{(2)})^T \\ \vdots  \\ (\Tnm{}^{(k)})^T  \end{array}        \right)
\bm{o}_{sv}' =
\left(       \begin{array}{c} \bm{c}^{(1)} \\ \bm{c}^{(2)} \\ \vdots  \\ \bm{c}^{(k)}  \end{array}        \right),
\end{equation}

where $\Tnm$ is our binary stretching transform matrix as defined as above. Therefore here $(\Tnm{}^{(k)})^T$ is an $M\! \times \! N$ binary matrix encoding which hr-pixels belong to each cell (i.e.\ element $m, n$ is 1 if pixel $m$ belongs to cell $n$, and 0 otherwise). $\bm{c}^{(k)}$ is a column vector of length $N$, element $n$ of which represents the sum of all the hr-pixel values in cell $n$ of sub-frame $k$. Conveniently, the vector $\bm{c}$ is already computed as part of the reconstruction of sub-frame $k$ (referring to Equation~\ref{eq:o-sv-recovery-hadamard}, we see that $\bm{c}=\frac{1}{N}\sum_n b_n \bm{h}_n$).

Note that, in the same way as $\Tnm{}$ maps from cell space to hr-pixel space, $\Tnm{}^T$ maps from hr-pixel space to cell space. These transformations are related by: 
$\Tnm{}^T \bm{A}^{-1} \Tnm{} = \bm{1}$, indicating that conversion from cell to hr-pixel space and back again is lossless. However, the reverse transformation $\bm{A}^{-1}\Tnm{} \Tnm{}^T$ does not equal $\bm{1}$, indicating that a transformation from hr-pixel to cell space and back again is \emph{not} lossless, and high resolution detail is lost in the transformation.

In practice, we solve for $\bm{o}_{sv}'$ using a least-squares method that is suitable for systems that may be locally overdetermined, critically determined, or underdetermined depending on the number of sub-frames available for the reconstruction. Our linear-constraint method can be sensitive to noise in the sub-frame measurements, and in particular noise is amplified in the highest spatial frequencies of the composite image (i.e.\ within the fovea). We suppress this noise by applying a spatially-variant smoothing constraint to the system of equations, which maintains generality as it is derived from the weighted average composite image formed from the raw data. This is achieved by adding extra rows to the system of equations incorporating the information present in the weighted average composite image. In this case the relative importance of the constraint terms can be tuned by solving for $\bm{o}_{sv}'$ using a weighted least-squares method. 

This effectively gives us a tunable compromise between noise suppression and faithful reproduction of the high spatial frequencies close to the cut-off frequency of the reconstruction: for example a greater weighting of the constraint leads to lower noise images but with high frequencies suppressed (non-uniformly across the field of view, reflecting the underlying measurements). In practical terms it is highly attractive to use the weighted average image as a constraint for the linear-constraint reconstruction, as it represents a ready-made space-variant noise suppression function, which would be non-trivial to otherwise synthesise from our irregular grids of sub-frame cells. Therefore the linear-constraint method is essentially equivalent to performing a deconvolution of the weighted-average reconstruction using the appropriate spatially-variant PSF.

In Equation~\ref{eq:linear-system} we can also account for cases where local motion has been detected in the images, as encoded in the difference map stacks. We do this simply by deleting any rows from the matrices $\Tnm^{(k)}$, along with the corresponding elements from vectors $\bm{c}^{(k)}$, that correspond to cells in sub-frame $k$ that are deemed to have changed.

We note that the matrices in the equations presented here are highly sparse, and an efficient implementation of the reconstruction code benefits significantly from exploiting this property. Note also that in the case of a regular square pixel grid (e.g. Fig.~\ref{fig1}(a)) the algebraic problem is separable in the $x$ and $y$ dimensions. This reduces the formal computational complexity of the problem, thus making it possible to implement a realtime reconstruction. However, it would be significantly more challenging to construct an irregular pixel grid (e.g. Fig.~\ref{fig1}(d)) that is separable in this way, while still meeting the other requirements for our geometry.\\

\noindent {\bf 6. Media file descriptions}.\\

\noindent {\bf Media 1: Real-time sub-frame display}. This movie shows data presented in Fig.~\ref{fig2}(a). The left hand panel shows the sub-frames captured at 8\,Hz (and processed and displayed in real-time). The super-sampling from one frame to the next can be seen both within the fovea where they repeat every 4 frames, and in the periphery where they repeat every 36 frames (the same as the length of the movie). The right hand panel shows the cell grid for each frame.\\

\noindent {\bf Media 2: Post-processed linear constraints reconstruction}. This movie shows data presented in Fig.~\ref{fig2}(c). The left hand panel shows the frame-by-frame linear constraints reconstruction. The high resolution appears to spread from the centre as in the periphery each new frame is fused with the existing data to improve the reconstruction. Right hand panel shows the effective exposure-time across the field-of-view. Initially the entire field-of-view has the same effective exposure-time as only a single frame has been recorded. In the centre only the most recent 4 sub-frames are used in the reconstruction (hence an effective exposure-time of 0.5\,s). Surrounding this data from progressively more frames back is used in the reconstruction (thus increasing the effective exposure time).\\

\noindent {\bf Media 3: Real-time motion tracking and fovea guidance}. This movie shows data presented in Fig.~\ref{fig3}(b). The top-left panel shows the low-resolution blip-frame (recorded after every 4th sub-frame in $\sim$\,31\,ms). The bottom-left panel shows the difference between the 2 most recent consecutive blip-frames. The region of the moving object is clearly visible. The blip-frame and blip-difference frame are reconstructed, analysed and displayed in real-time. The middle panel shows the sub-frames captured at 8\,Hz (and processed and displayed in real-time). Here the fovea is programmed to follows the moving part of the scene to image it at high resolution. Additionally, 20\% of the time the fovea is programmed to jump to a random location within the field-of-view that was not recently accessed. This is performed to ensure that all of the sub-frames are at least intermittently sampled, improving the quality of the longer exposure super-sampled reconstruction of static parts of the scene (shown in Media 4). The global position of the fovea is updated at 2\,Hz (every 4 sub-frames), based on the blip-difference frame analysis. During each 4 sub-frame fixation phase in-between blip-frame measurements, the fovea cell footprints are shifted by 4 half-cell displacements in $x$ and $y$~\cite{Sun2016}. This enables the resolution within the fovea to be doubled by combining the 4 measurements should the scene with the fovea remain static for the time of the measurements. The right hand panel shows the cell grid for each frame.\\

\noindent {\bf Media 4: Real-time weighted-averaging and post-processed linear constraints reconstruction of a dynamic scene}. This movie shows data presented in Fig.~\ref{fig3}(c-d). The top-left panel shows the post-processed linear-constraints reconstruction of the raw data shown in Media 3. The effective exposure-time of this reconstruction is shown in the top-right panel. Here the minimum effective exposure-time is 0.25\,s and the maximum is 4\,s. Therefore in the parts of the scene that are currently deemed to be moving, we display the average of the most recent 2 frames. In other regions the data used in this reconstruction is flushed after a maximum of 4\,s (i.e.\ the maximum number of previous sub-frames from which data is used is 32). In this case as in some regions no change was detected throughout the entire duration of the clip (15\,s, $~\sim$\,120 sub-frames), the maximum effective-exposure could have been 15\,s. We show the case for a reduced 4\,s maximum effective exposure-time to demonstrate the super-sampled recovery of the entire field-of-view.

The bottom-left panel shows the real-time weighted-averaging reconstruction. Here instead of choosing a hard limit on the maximum effective exposure-time we have weighted older frames less prominently, which protea recent measurements.

The bottom-right panel shows a uniform resolution video of a similar scene, using the same measurement resource as the data in the other panels. Here the data is completely flushed every frame (0.125\,s), however the resolution is never high enough to identify the lettering on the moving sign, or any of the features on the resolution target in the background.

\end{document}